\documentclass[fullpaper,final]{nldl}
\paperID{TBC}%{35} %{TBC}
\vol{V}
\usepackage{mathtools}

\usepackage{graphicx}

\usepackage{booktabs}

\usepackage{enumitem}

\usepackage{algorithm}
\usepackage{algorithmic}

\usepackage{listings}
\usepackage{biblatex} % added by asra
\usepackage{hyperref}
\usepackage{url}
\hypersetup{
  pdfusetitle,
  colorlinks,
  linkcolor = BrickRed,
  citecolor = NavyBlue,
  urlcolor  = Magenta!80!black,
}

\title{Predicting Future Organ Dysfunction in ICU Patients Using Temporal Convolutional Networks on MIMIC-IV Data}
\author[]{Razan Albouq and Asra Aslam, University of Sheffield, UK, \texttt{\{ralbouq1, a.aslam\}@sheffield.ac.uk} }
\begin{document}
\maketitle

\begin{abstract}
Predicting future organ dysfunction in Intensive Care Unit (ICU) patients is critical for early clinical intervention, yet existing machine learning approaches have largely treated the Sequential Organ Failure Assessment (SOFA) score as an input to binary mortality prediction rather than as a continuous clinical outcome in its own right. We investigate the extent to which a Temporal Convolutional Network (TCN) can predict next-day SOFA scores from multivariate ICU time-series data extracted from MIMIC-IV, characterise the relative contribution of each organ system to total SOFA variance and deterioration, and identify distinct trajectory patterns across ICU stays. A residual TCN trained on three-day sliding windows achieved a five-fold cross-validation $R^2$ of $0.740 \pm 0.013$ and MAE of $1.431 \pm 0.022$, outperforming a na\"ive persistence baseline on RMSE and $R^2$. SHAP interpretability analysis revealed that the model functions primarily as a severity-anchoring mechanism rather than a true sequence model, with predictions dominated almost entirely by the most recent observation day. Cardiovascular dysfunction emerged as the strongest discriminator of both cross-sectional severity and acute deterioration, and unsupervised trajectory clustering identified two clinically meaningful phenotypes, an improving group (58.9\%) and a persistently severe group (41.1\%), differentiated by cardiovascular, hepatic, coagulation, and renal involvement. We conclude that TCNs can extract meaningful predictive signal from ICU physiological data, but that short input windows and complete-case selection bias currently limit their clinical utility, motivating future work on longer input horizons, alternative missing-data strategies, and external validation.
\end{abstract}

\section{Introduction}
\label{sec:intro}
Intensive Care Units (ICUs) require continuous monitoring of critically ill patients, and early recognition of individuals at risk of organ failure enables clinical prioritisation and intervention, reducing the risk of sepsis, multi-organ failure, and mortality~\citep{singer2016}. Rising demand, limited resources, and workforce shortages continue to place pressure on the ability of health systems to deliver timely, equitable critical care~\citep{nhsengland2023}, sharpening the need for tools that can anticipate deterioration before it becomes clinically obvious.

A central clinical measure of organ dysfunction in ICU settings is the Sequential Organ Failure Assessment (SOFA) score, which evaluates six organ systems, respiratory, coagulation, hepatic, cardiovascular, neurological, and renal, assigning each a subscore from 0 to 4 based on physiological measurements, yielding a total score between 0 and 24 that correlates with mortality risk~\citep{vincent1996}. The derivation of SOFA scores relies on ICU electronic health record (EHR) data, which are highly heterogeneous, encompassing vital signs, laboratory tests, and vasopressor infusion rates recorded at varying measurement frequencies across organ systems. This heterogeneity imposes substantial constraints on temporal modelling and necessitates careful methodological decisions at every stage of model development.

Prior machine learning (ML) work in this space has largely used SOFA as an \emph{input feature} to binary mortality-prediction models rather than forecasting organ dysfunction as a continuous outcome in its own right (Section~\ref{sec:related}). This leaves open the question of how accurately future SOFA trajectories can be predicted directly from routinely collected ICU data, which organ systems drive overall severity and acute deterioration, and whether distinct clinical phenotypes exist within ICU stay trajectories.

This paper addresses three research questions using the MIMIC-IV database~\citep{johnson2023}:
\begin{itemize}%[leftmargin=*]
  \item \textbf{RQ1}: How accurately can a Temporal Convolutional Network (TCN) predict future SOFA scores at a short-term (24-hour) horizon from multivariate ICU time-series data?
  \item \textbf{RQ2}: Which organ system components of the SOFA score contribute most to total score variance and deterioration events?
  \item \textbf{RQ3}: What distinct SOFA trajectory patterns exist among ICU stays, and how do they differ in organ-system involvement and severity?
\end{itemize}

Our contribution is threefold. First, we predict continuous SOFA trajectories across a general ICU population using a residual TCN, rather than treating SOFA purely as a mortality-prediction feature. Second, we characterise the relative contribution of each organ system to total SOFA variance and to acute deterioration, using two independent statistical measures. Third, we identify distinct ICU stay subgroups from SOFA temporal patterns using unsupervised clustering, and show that these subgroups are governed by the same organ systems identified in the component analysis, a convergence between two independent analytical approaches that strengthens confidence in the underlying clinical pattern.

\section{Related Work}
\label{sec:related}

\textbf{ML in critical care.} Deep learning has been increasingly applied to ICU outcome prediction to assist clinicians in making timely, data-driven decisions. \citep{wei2025} used dynamic SOFA component scores from the first four ICU days to predict post-discharge mortality (28-day, 90-day, and 1-year) among sepsis survivors. \citep{liu2022} developed a time-incorporated, SOFA-based XGBoost model for in-ICU mortality that captured the duration and trajectory of organ dysfunction across six consecutive SOFA measurements within the first 72 hours of admission. \citep{lim2024} took a broader approach, combining deep learning and gradient boosting on routinely collected vitals and laboratory results that update in real time to predict short-term mortality. These studies collectively establish the applicability of ML to ICU outcome prediction using routinely collected physiological data, but each treats the SOFA score as an input to binary mortality prediction rather than as a continuous clinical outcome in its own right, leaving the direct forecasting of organ dysfunction severity unaddressed.

\textbf{Predicting and analysing the SOFA metric.} \citep{montomoli2021} applied the XGBoost algorithm to COVID-19 ICU patients to predict the five-day delta in SOFA score as a binary outcome indicating whether organ dysfunction would worsen or improve. \citep{asuroglu2021} approached SOFA prediction as a continuous regression task, combining CNN-extracted features with a Random Forest regressor to predict exact SOFA score values in sepsis patients using seven bedside vital signs from MIMIC-III. Both studies relied on non-sequential architectures, and neither captured organ dysfunction as a continuous temporal trajectory across a general ICU population, nor examined how individual organ system components contribute to total SOFA variance and deterioration dynamics.

\textbf{TCNs for clinical time-series prediction.} Recurrent architectures such as LSTM and GRU networks process sequences step-by-step, making efficient parallelisation difficult and rendering them susceptible to vanishing gradients over long sequences~\citep{bai2018}. In ICU settings specifically, where longitudinal variables are measured at varying frequencies and missing values are common, recurrent networks have been shown to struggle with the irregular sampling characteristic of clinical data~\citep{catling2020}. TCNs address these limitations via dilated causal convolutions, which capture long-range temporal dependencies while supporting fully parallelised training; the causal constraint ensures predictions at each time step are conditioned only on past observations, while exponentially increasing dilation rates expand the receptive field without proportionally increasing model depth~\citep{bai2018}. \citep{bednarski2022} applied TCNs to ICU time-series data for mortality and length-of-stay prediction, finding that TCNs outperformed GRU and random forest baselines. These properties motivate TCNs as a strong candidate architecture for modelling continuous SOFA trajectories, where capturing temporal dependencies across organ systems is central to accurate next-day prediction.

\textbf{Handling missing ICU data.} Temporal EHR data typically involve substantial missing values due to the infeasibility of continuous measurement in clinical settings. \citep{singh2021} demonstrated that missingness in critical care EHR data is informative rather than random, and that ignoring missingness patterns can reduce ML model performance on ICU data. \citep{afkanpour2024} concluded through a systematic review that selecting the most appropriate imputation method requires careful consideration of the missingness mechanism, pattern, and ratio specific to the dataset, rather than applying a default technique. Even when imputation is attempted with this consideration, \citep{kazijevs2023} demonstrated through benchmarking across five healthcare datasets that no single method consistently outperforms others, with performance depending critically on data type, missing-value rate, and missingness mechanism. Where imputation therefore offers no clear methodological advantage, \citep{websterclark2024} demonstrated that complete-case analysis can produce unbiased estimates under specific epidemiological conditions, providing a theoretically grounded alternative when the missingness mechanism is complex and poorly characterised.

\section{Methods}
\label{sec:methods}

\subsection{Data Collection}

This study used MIMIC-IV, a publicly available, de-identified collection of EHRs from over 76,000 ICU admissions at the Beth Israel Deaconess Medical Centre between 2008 and 2019~\citep{johnson2023}. Variables required to compute all six SOFA organ system components were extracted from the \texttt{chartevents}, \texttt{labevents}, and \texttt{inputevents} tables, including vital signs, laboratory values, and vasopressor infusion rates (see Appendix~\ref{app:variables}, Table~\ref{tab:variables}, for the full extraction reference).

\subsection{Data Preprocessing}
\label{sec:preprocessing}

Events were merged with ICU stay metadata to compute a relative ICU day variable, with day zero representing the admission day, and only measurements recorded within the documented ICU stay period were retained. Each variable was subjected to physiological range cleaning to remove erroneous values, with bounds defined according to established clinical norms for each measurement type.

Vasopressor infusion rates were standardised to micrograms per kilogram per minute for consistent SOFA cardiovascular scoring, with drug-specific physiological caps applied following clinical guidelines. Since PaO$_2$ and FiO$_2$ are measured independently and at different frequencies, each PaO$_2$ reading was paired with the nearest FiO$_2$ value charted on the same ICU stay using a nearest-neighbour temporal merge, and the lowest paired P/F ratio of each ICU day was retained for SOFA scoring.

All variables were aggregated into daily summaries reporting the worst observed value for each day: MAP, GCS, and platelet count were summarised using the daily minimum; creatinine and bilirubin using the daily maximum; and vasopressor doses using the daily maximum rate per drug, with zero assigned to patient-days on which no vasopressor was administered.

A complete-case analysis was adopted, retaining only patient-days on which all six SOFA components were directly observed, supported by \citep{websterclark2024}, who demonstrated that complete-case analysis can produce unbiased estimates under specific epidemiological conditions. This decision was necessary because ICU variables have highly heterogeneous measurement frequencies: vital signs such as MAP and GCS are charted continuously, while laboratory tests including creatinine, bilirubin, platelets, and PaO$_2$ are ordered only one to three times per week in routine clinical practice, making forward-fill, backward-fill, and linear interpolation unreliable across such multi-day gaps. This approach is consistent with guidance from \citep{afkanpour2024}, but is also acknowledged as a limitation, since it introduces a selection bias towards the most intensively monitored, and hence more severely ill, patients (Section~\ref{sec:discussion}). Patients under 18 years of age were excluded, as the SOFA score was developed and validated for use in adult ICU populations~\citep{vincent1996}.

The six SOFA subscores were computed per ICU stay day using the validated clinical thresholds established by \citep{vincent1996}, and the total SOFA score was computed as the arithmetic sum of all six subscores, ranging from 0 to 24. Daily data were sorted by ICU stay and day, and overlapping sliding windows of three consecutive days were constructed as model inputs, with the SOFA score on the immediately following day serving as the prediction target. This window length reflects the standard clinical assessment frequency of the SOFA score, calculated at 24 hours after ICU admission and every 48 hours thereafter~\citep{vincent1998}. ICU stay IDs were partitioned into training (80\%) and test (20\%) sets prior to sequence construction, ensuring that no ICU stay appeared in both sets and preventing temporal data leakage. Z-score normalisation was fitted exclusively on the training set and subsequently applied to both sets.

\subsection{Model Architecture}
\label{sec:architecture}

A TCN was implemented following the residual architecture described by \citep{bai2018}, with LayerNormalization substituted for weight normalisation to ensure TensorFlow/Keras compatibility~\citep{ba2016}. The model comprised four stacked residual blocks with exponentially increasing dilation rates of 1, 2, 4, and 8, each block containing two dilated causal convolutional layers with a kernel size of 3, followed by LayerNormalization, ReLU activation, and dropout (rate = 0.2), with a $1{\times}1$ skip connection from block input to output. This residual design allows the network to learn corrections to its input rather than the full mapping, while the exponentially increasing dilation rates expand the receptive field to capture temporal patterns at multiple scales, improving gradient flow and avoiding the vanishing-gradient issues associated with recurrent networks~\citep{bai2018}. The output of the final residual block was passed through a global average pooling layer, a fully connected layer of 32 units with ReLU activation, and a single linear output neuron producing the continuous SOFA prediction. All convolutional layers used 64 filters.

\subsection{Training Strategy}

The model was compiled using the Adam optimiser~\citep{kingma2014}, selected for its adaptive learning rate properties and robust performance across deep learning tasks, with an initial learning rate of 0.001. Mean squared error (MSE) was used as the loss function, consistent with standard practice in continuous regression tasks where penalising larger prediction errors is desirable. To improve model sensitivity to clinically significant deterioration, sample weights proportional to SOFA severity were applied during training, defined as $\text{weight} = 1 + \text{SOFA}/24$. Training ran for a maximum of 60 epochs with a batch size of 64. Early stopping was applied with a patience of 8 epochs and restoration of best weights, and a learning-rate scheduler reduced the learning rate by a factor of 0.5 when validation loss failed to improve for four consecutive epochs. All random seeds were fixed prior to training to ensure reproducibility.

\subsection{Model Interpretability}
\label{sec:interpretability}

SHAP (SHapley Additive exPlanations) values were computed to identify the most influential physiological features contributing to each SOFA prediction~\citep{lundberg2017}. The SHAP KernelExplainer was used in place of gradient-based SHAP methods (DeepExplainer, GradientExplainer) due to compatibility constraints with the TensorFlow version used in this study. KernelExplainer is a model-agnostic approach that computes SHAP values via weighted linear regression over feature coalitions~\citep{lundberg2017}, which can make larger explanation samples computationally prohibitive. A background sample of 50 training sequences and an explanation sample of 50 test sequences were used to balance explanation quality against computational cost. SHAP values were averaged across the three time steps of each input window to produce a single mean absolute importance score per feature, and a temporal heatmap was produced to visualise how feature importance varied across the three input days.

\subsection{SOFA Component Contribution} % Analysis}
\label{sec:component}

To address RQ2, the relative contribution of each of the six SOFA organ system subscores to total score variance was analysed. The variance of each subscore was expressed as a percentage of total SOFA variance, and Pearson correlations between each subscore and the total score were computed to identify the most discriminating components. Mean subscore profiles were then compared across the three SOFA severity bands to assess how organ system involvement changes with clinical severity. Finally, mean subscores on clinically stable days ($|\Delta\text{SOFA}| < 2$) were compared with those on deteriorating days ($\Delta\text{SOFA} \geq 2$) to quantify which organ systems show the greatest elevation during deterioration events.

\subsection{SOFA Trajectory Clustering}
\label{sec:clustering}

To address RQ3, an unsupervised clustering analysis was conducted on ICU stay SOFA trajectories to identify distinct temporal patterns of organ dysfunction. Each ICU stay's SOFA time series was interpolated onto a fixed seven-point normalised timeline using linear interpolation, enabling comparison across ICU stays of different lengths; only ICU stays with at least two observed SOFA days were included. K-means clustering was applied to the resulting trajectory matrix~\citep{macqueen1967}, with the optimal number of clusters determined by silhouette score analysis across $k$ values from 2 to 7~\citep{rousseeuw1987}. Each resulting cluster was profiled using a radar chart of its centroid trajectory shape, mean SOFA total, and mean subscore composition, and clusters were assigned descriptive clinical labels based on trajectory direction and mean severity level.

%\subsection{Ethical Considerations}
%
%This project used secondary data from the MIMIC-IV database, which is fully de-identified and publicly accessible through the PhysioNet credentialed user agreement~\citep{johnson2023}. Access was granted upon completion of the Collaborative Institutional Training Initiative (CITI) course on human research and acceptance of the PhysioNet data use agreement. No direct patient interaction or identifiable information was involved at any stage of the analysis, and the project posed no additional ethical risks beyond the ethical oversight already governing the MIMIC-IV database.

\section{Performance Evaluation} %{Empirical Results and Evaluation}
\label{sec:results}

\subsection{Experimental Setup and Metrics} %{Model Evaluation}
\label{sec:evaluation}

Model performance was assessed using three complementary regression metrics: mean absolute error (MAE), root mean squared error (RMSE), and the coefficient of determination ($R^2$), consistent with standard practice in regression modelling~\citep{chai2014}. These metrics collectively assess average prediction deviation, sensitivity to large errors, and proportion of variance explained. A na\"ive persistence baseline was computed by predicting that tomorrow's SOFA score equals today's, providing a clinically meaningful reference point for evaluating model performance. Results were additionally stratified by SOFA severity band (Low: 0--5, Moderate: 6--10, Severe: $\geq$11) to assess whether model accuracy varied across clinically distinct ICU stay subgroups. A binary deterioration-detection analysis was conducted by applying a threshold of $\Delta\text{SOFA} \geq 2$ to both predictions and actual values, reporting precision, recall, F1-score, and AUC-ROC to evaluate the model's ability to identify acutely deteriorating patients.

Model generalisation was assessed through five-fold ICU-stay-level cross-validation, in which all sequences from each ICU stay were assigned to the same fold to preserve temporal integrity and prevent ICU-stay-level data leakage. Each fold trained a fresh model with fold-specific random seeds. Cross-validation results are reported as mean $\pm$ standard deviation across the five folds and constitute the primary performance figures of this study, as recommended by~\citep{kohavi1995}.

%\paragraph{Ethical Considerations} 
\noindent\textit{Ethical Considerations:} This research used secondary data from the MIMIC-IV database, which is fully de-identified and publicly accessible through the PhysioNet credentialed user agreement~\citep{johnson2023}. Access was granted upon completion of the Collaborative Institutional Training Initiative (CITI) course on human research and acceptance of the PhysioNet data use agreement. 

%No direct patient interaction or identifiable information was involved at any stage of the analysis, and the project posed no additional ethical risks beyond the ethical oversight already governing the MIMIC-IV database.

\subsection{Regression Performance}
\label{sec:regression-results}

\begin{figure*} %[tb]
	\centering
	\includegraphics[width=\linewidth]{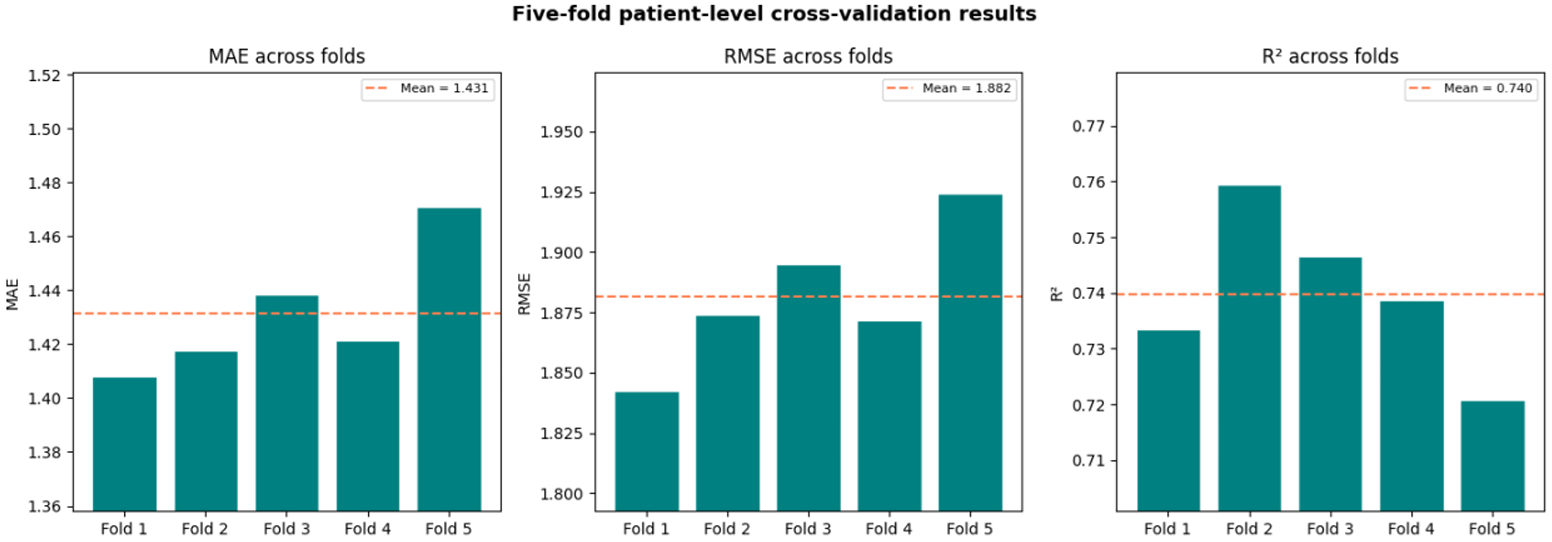}
	\caption{MAE, RMSE, and $R^2$ across five cross-validation folds, with mean performance shown as a dashed line.}
	\label{fig:cv}
\end{figure*}

On the holdout test set, comprising 4,914 test sequences, the model achieved an MAE of 1.462, an RMSE of 1.906, and an $R^2$ of 0.715. Across the five cross-validation folds, the model achieved a mean MAE of $1.431 \pm 0.022$, RMSE of $1.882 \pm 0.027$, and $R^2$ of $0.740 \pm 0.013$ (Table~\ref{tab:performance}, Figure~\ref{fig:cv}). Individual fold $R^2$ values ranged narrowly from 0.721 to 0.759, reflecting stable generalisation across different ICU stay subsets, and the low standard deviation across all three metrics confirms that model performance was consistent across random ICU stay partitions~\citep{kohavi1995}.

\begin{table} %[tb]
	\centering
	\caption{TCN performance across five-fold ICU stay-level cross-validation, single holdout evaluation, and the na\"ive persistence baseline.}
	\label{tab:performance}
	\resizebox{\linewidth}{!}{%
		\begin{tabular}{lccc}
			\toprule
			& MAE & RMSE & $R^2$ \\
			\midrule
			CV mean $\pm$ SD & $1.431 \pm 0.022$ & $1.882 \pm 0.027$ & $0.740 \pm 0.013$ \\
			Holdout & 1.462 & 1.906 & 0.715 \\
			Persistence baseline & 1.401 & 1.933 & 0.707 \\
			\bottomrule
	\end{tabular}}
\end{table}

\begin{figure*} %[tb]
	\centering
	\includegraphics[width=\linewidth]{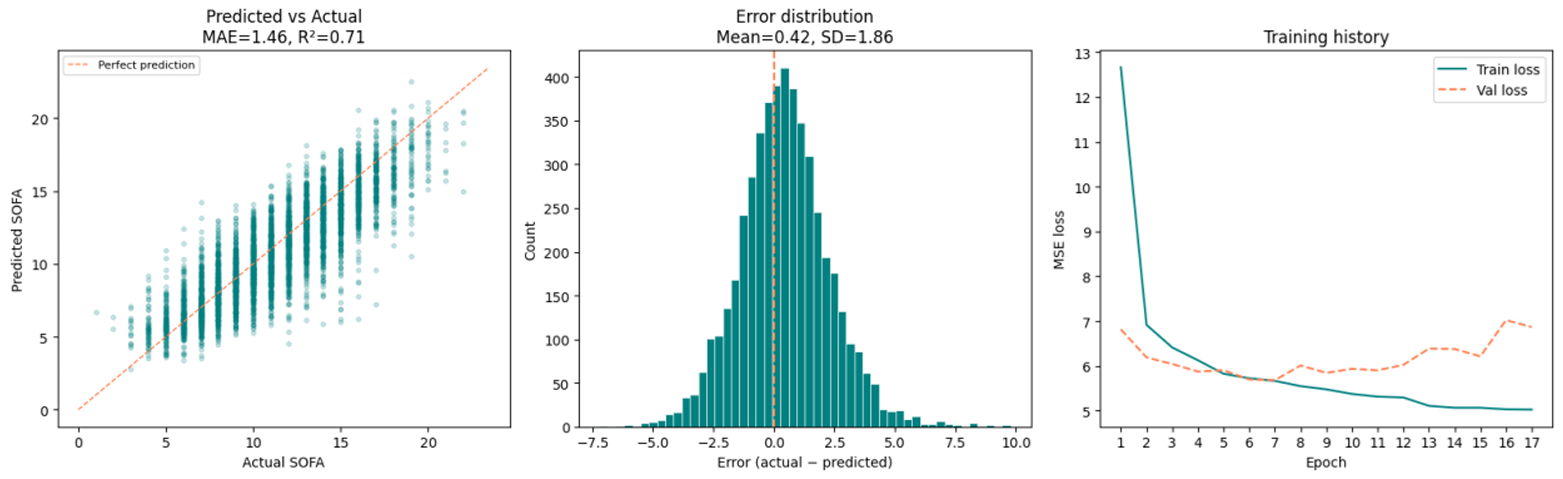}
	\caption{Predicted vs actual SOFA Evaluation (left), error distribution (centre), and training history (right).}
	\label{fig:eval}
\end{figure*}

The predicted-versus-actual scatter plot showed a strong linear relationship across the full SOFA range, with predictions closely aligned with the diagonal reference line, though a banding pattern suggests the model tends to predict rounded SOFA values (Figure~\ref{fig:eval}). The error distribution had a mean of 0.42 and a standard deviation of 1.86, indicating that the model tends to under-predict actual SOFA scores on average. Training showed rapid initial convergence, with validation loss reaching its minimum at epoch 7 before rising, triggering early stopping at epoch 17 with best weights restored from epoch 7.

\begin{figure*}[h] %[tb]
	\centering
	\includegraphics[width=0.88\linewidth]{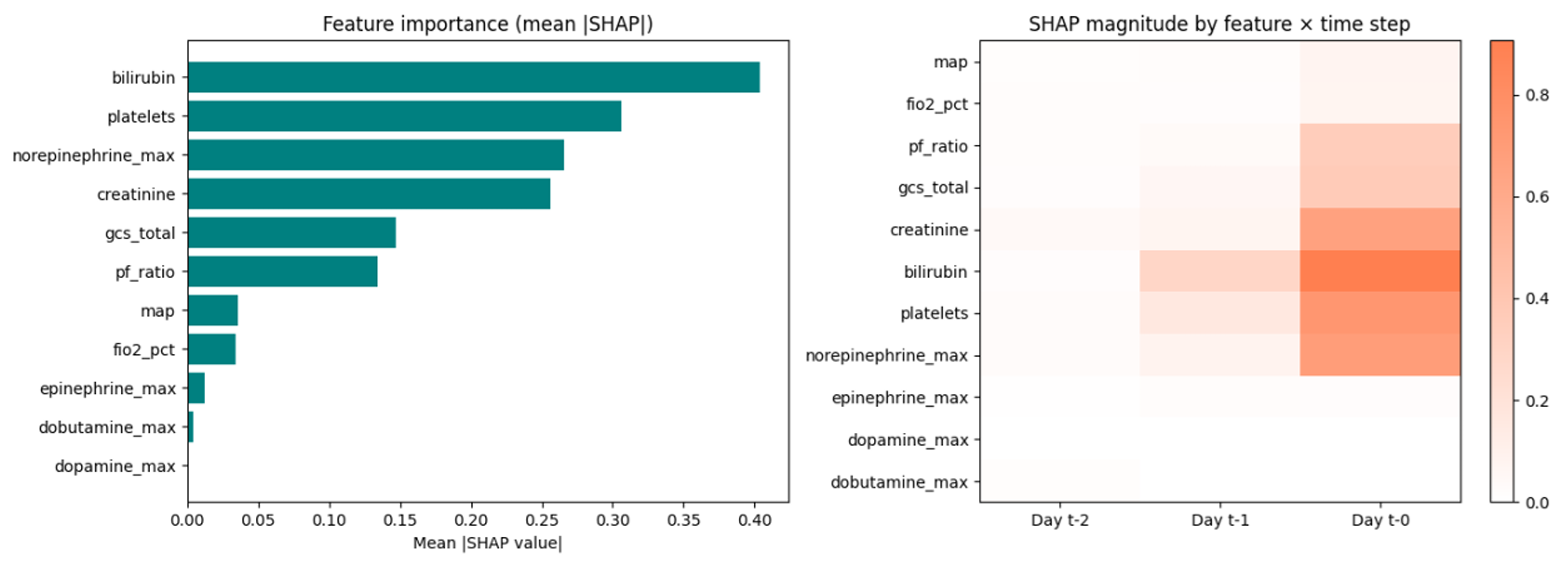}
	\caption{SHAP feature importance (left) and temporal SHAP heatmap by feature and time step (right).}
	\label{fig:shap}
\end{figure*}
The na\"ive persistence baseline achieved an MAE of 1.401, RMSE of 1.933, and $R^2$ of 0.707 on the holdout test set (Table~\ref{tab:performance}). The TCN outperformed the baseline on RMSE (0.027 improvement) and $R^2$ (0.008 improvement), demonstrating that the model extracts a genuine temporal signal beyond simple score persistence. The exception was MAE, where the TCN's score of 1.462 was marginally higher than the baseline's 1.401. Since RMSE penalises larger errors quadratically while MAE weights all errors equally, this pattern indicates the TCN reduces clinically significant large errors at a small cost to average accuracy. The baseline $R^2$ of 0.707 confirms that SOFA scores are strongly autocorrelated from daily, making persistence a competitive reference point.

%day to day 
Performance varied substantially across SOFA severity bands (Appendix~\ref{app:extra}, Table~\ref{tab:severity}). In the Severe band (SOFA $\geq$ 11, $n=2{,}739$), the model achieved the only positive $R^2$ across the three bands (0.134); the Moderate (6--10, $n=1{,}967$) and Low (0--5, $n=208$) bands showed negative $R^2$ ($-0.363$ and $-6.256$ respectively). These stratified results are directly attributable to the SOFA score distribution in the dataset, which peaks between 8 and 11 due to the complete-case selection mechanism: patients with all six SOFA components measured simultaneously are predominantly those receiving the most intensive monitoring, who are by definition moderately to severely ill, resulting in a training distribution concentrated around moderate-to-severe scores. The negative $R^2$ in the Low and Moderate bands does not indicate that the model fails clinically in those ranges, but rather that its predictions are better calibrated to the severity range where the majority of the training data lie.

\subsection{Deterioration Detection}

When regression predictions were thresholded to answer the binary clinical question of whether SOFA would increase by $\geq 2$ points in the next 24 hours, the model's outputs reflected the substantial class imbalance in the underlying data (Table~\ref{tab:deterioration}, Appendix~\ref{app:extra}). Of 733 actual deteriorating sequences, the model correctly identified only 19 (recall $= 0.03$), while 4,147 of 4,181 stable sequences were correctly identified (precision $= 0.36$ for the deteriorating class), yielding an F1-score of 0.05 for the deteriorating class and an overall accuracy of 0.85 driven by correct identification of stable sequences. Despite the low recall, the AUC-ROC of 0.663 indicates that the model's continuous predictions contain meaningful discriminative signal for deterioration risk, though the binary decision threshold is poorly calibrated for detecting the minority deteriorating class. This reflects two compounding factors: the class imbalance between stable and deteriorating sequences heavily incentivises the model to predict stability, and the complete-case cohort under-represents the dynamic, fluctuating patients most likely to deteriorate, since patients with rapidly changing physiology are less likely to have all six SOFA components simultaneously observed on consecutive days.

Individual ICU stay trajectory plots (Appendix~\ref{app:extra}, Figure~\ref{fig:trajectories}) revealed that admissions with stable or gradually changing trajectories were tracked most closely, while performance degraded systematically as the rate and magnitude of SOFA change increased. In the most volatile sampled admission, in which SOFA oscillated between approximately 8 and 16 over roughly 28 ICU days, deterioration peaks were underestimated by approximately 2 to 3 points; in another, SOFA dropped sharply to approximately 4 before recovering to approximately 9, yet predictions remained flat near 5--6 throughout, failing to capture either transition. These patterns reflect a consistent smoothing bias: the model approximates gradual trends reliably but cannot anticipate abrupt directional changes, a behaviour consistent with its strong dependence on the most recent observation day, as the SHAP analysis below confirms.

\subsection{SHAP Feature Importance}
\label{sec:shap-results}

The SHAP analysis revealed that the model's predictions were driven primarily by hepatic, coagulation, renal, and cardiovascular dysfunction, with bilirubin, platelets, norepinephrine dose, and creatinine collectively accounting for the dominant share of predictive signal, carrying mean absolute SHAP values of 0.403, 0.306, 0.266, and 0.256 respectively (Figure~\ref{fig:shap}). GCS total and P/F ratio contributed moderately (0.147 and 0.130), while MAP and FiO$_2$ contributed negligibly (below 0.05). The near-zero importance of dopamine, dobutamine, and epinephrine reflects their low prevalence in the cohort relative to norepinephrine, consistent with its designation as first-line vasopressor therapy in sepsis management.

Critically, the temporal heatmap in Figure~\ref{fig:shap} exposes a more fundamental characteristic: SHAP magnitudes were concentrated almost entirely in Day $t{-}0$ for all features, with Day $t{-}1$ carrying only a secondary signal for the four highest-importance features and Day $t{-}2$ contributing negligibly across all features. This reveals a gap between architectural intent and learned behaviour: the TCN was designed to exploit long-range temporal dependencies through dilated causal convolutions, yet the model effectively learned to ignore the directional information contained in the earlier time steps, anchoring predictions in the most recent physiological snapshot rather than the trajectory leading to it. This behaviour is a direct consequence of the strong day-to-day autocorrelation in SOFA scores: since today's score already explains the majority of tomorrow's variance, the model found no sufficient training incentive to exploit the earlier window days, reducing a sequence model to a severity-anchoring mechanism in practice.

\subsection{SOFA Component Contribution}
\label{sec:component-results}
\begin{figure*}[h] %[tb]
	\centering
	\includegraphics[width=0.9\linewidth]{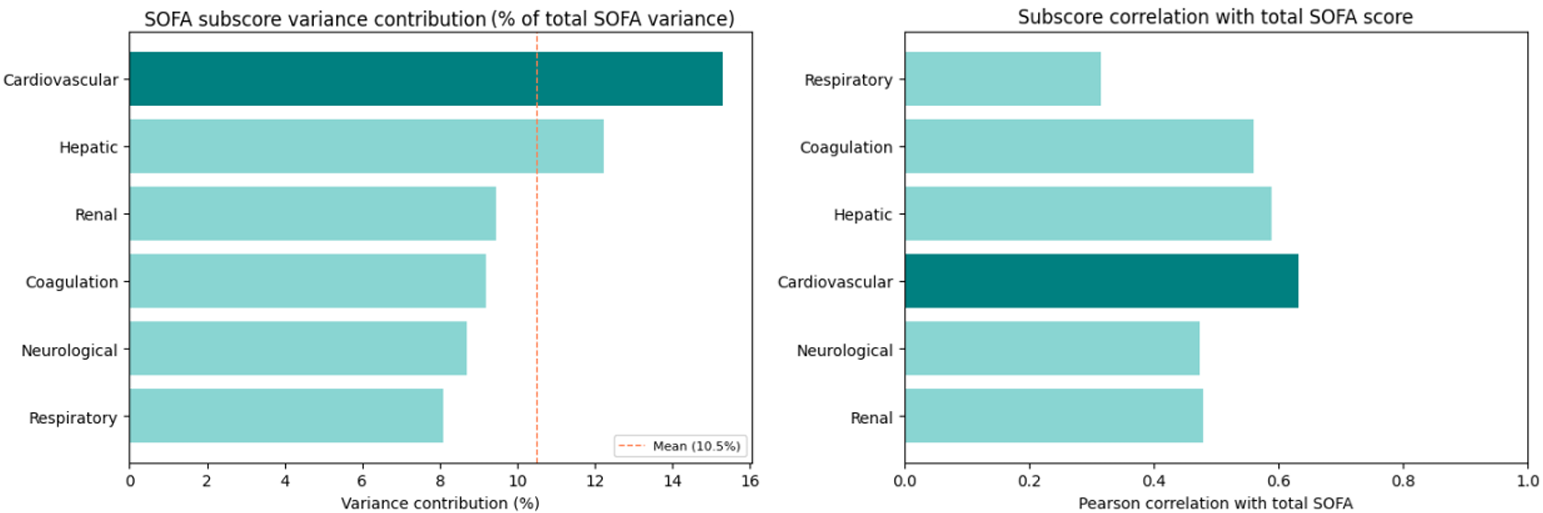}
	\caption{SOFA subscore variance contribution (left) and Pearson correlation with total SOFA score (right).}
	\label{fig:variance}
\end{figure*}
\begin{figure*}[h] %[tb]
	\centering
	\includegraphics[width=0.9\linewidth]{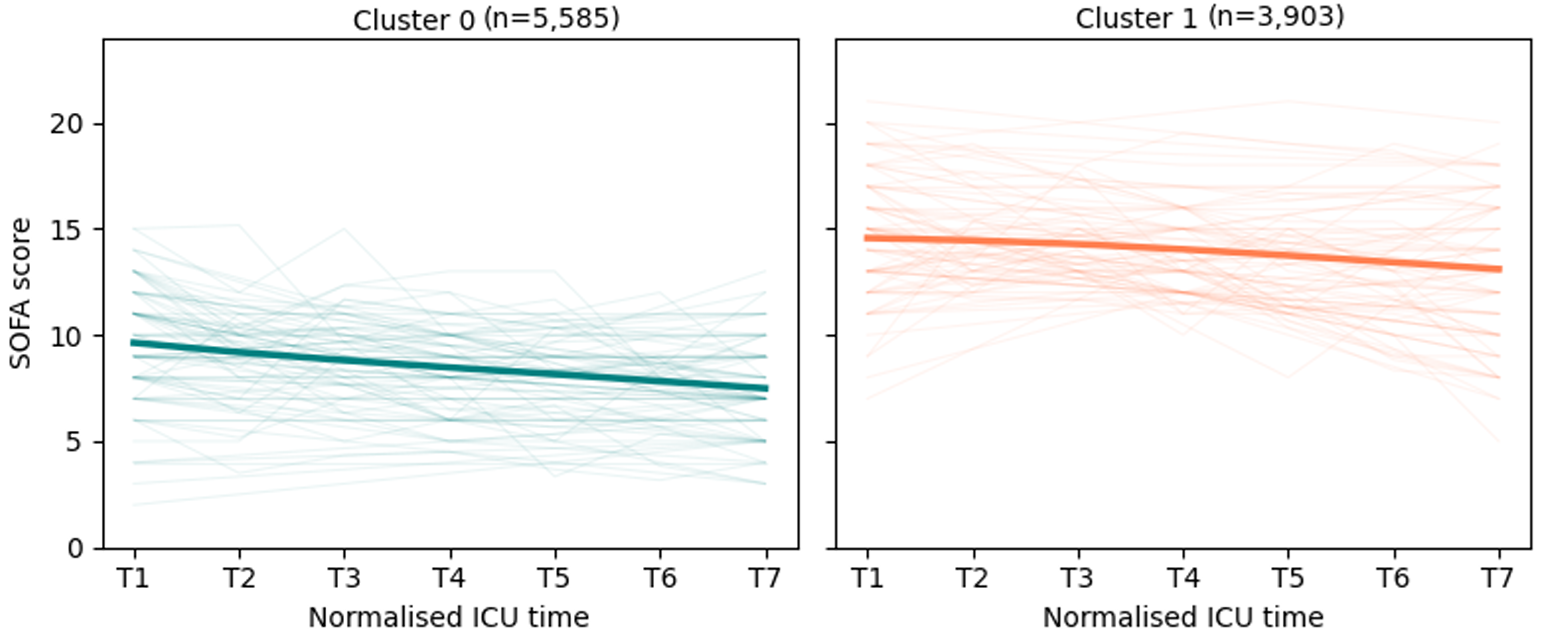}
	\caption{SOFA trajectory clusters: Cluster 0 (improving, left) and Cluster 1 (persistently severe, right).}
	\label{fig:clusters}
\end{figure*}
The six SOFA organ system subscores exhibited markedly different distributions across the complete-case cohort (Appendix~\ref{app:extra}, Figure~\ref{fig:boxplot}). Neurological and respiratory dysfunction were the most consistently present, with means of $2.999 \pm 1.114$ and $2.770 \pm 1.074$, indicating that moderate-to-severe impairment in these systems was the norm in this cohort. In contrast, hepatic, coagulation, and renal subscores were zero-inflated, with medians at or near 0 despite means of $1.008 \pm 1.321$, $0.965 \pm 1.144$, and $1.122 \pm 1.160$ respectively, reflecting dysfunction concentrated in a smaller subset of severely ill patients. Cardiovascular dysfunction occupied an intermediate position, with a mean of $1.961 \pm 1.477$ and the widest spread across the six subscores.

Both variance contribution and Pearson correlation independently identified the same four organ systems as the most discriminating components of total SOFA severity (Figure~\ref{fig:variance}): cardiovascular ranked first on both measures (15.3\% variance, $r=0.63$), followed by hepatic (12.2\%, $r=0.59$), renal (9.4\%, $r=0.48$), and coagulation (9.2\%, $r=0.56$). Neurological and respiratory ranked lowest on both measures (8.7\%/$r=0.48$ and 8.1\%/$r=0.32$) \emph{despite} having the highest mean subscores, because their consistent elevation across the cohort left little discriminating variation between ICU stays. The dominance of cardiovascular dysfunction reflects binary nature of vasopressor requirement, which creates the widest spread between patient-days of haemodynamic stability and severe circulatory failure.

All six organ systems increased monotonically with overall SOFA severity, but the rate of escalation differed substantially across systems (Appendix~\ref{app:extra}, Figure~\ref{fig:severity}): cardiovascular dysfunction rose from 0.574 in the Low band to 2.755 in the Severe band, a nearly fivefold escalation, while hepatic, coagulation, and renal subscores also increased steeply. Neurological and respiratory subscores were already elevated in low-severity patients, leaving comparatively less room for escalation, consistent with their low variance contribution above.

For acute deterioration (Appendix~\ref{app:extra}, Figure~\ref{fig:deterioration}), cardiovascular dysfunction showed largest mean subscore elevation on deteriorating versus stable days ($+0.546$), followed by respiratory ($+0.398$), neurological ($+0.296$), renal ($+0.271$), coagulation ($+0.197$), and hepatic ($+0.148$), based on 5,756 deteriorating days and 22,673 stable days. Notably, respiratory ranked second on acute deterioration involvement despite ranking last on variance contribution and correlation, suggesting that while respiratory dysfunction contributes little to between-patient variation in severity, it fluctuates acutely during deterioration events even when chronically elevated.

\begin{figure}
	\centering
	\includegraphics[width=0.9\linewidth]{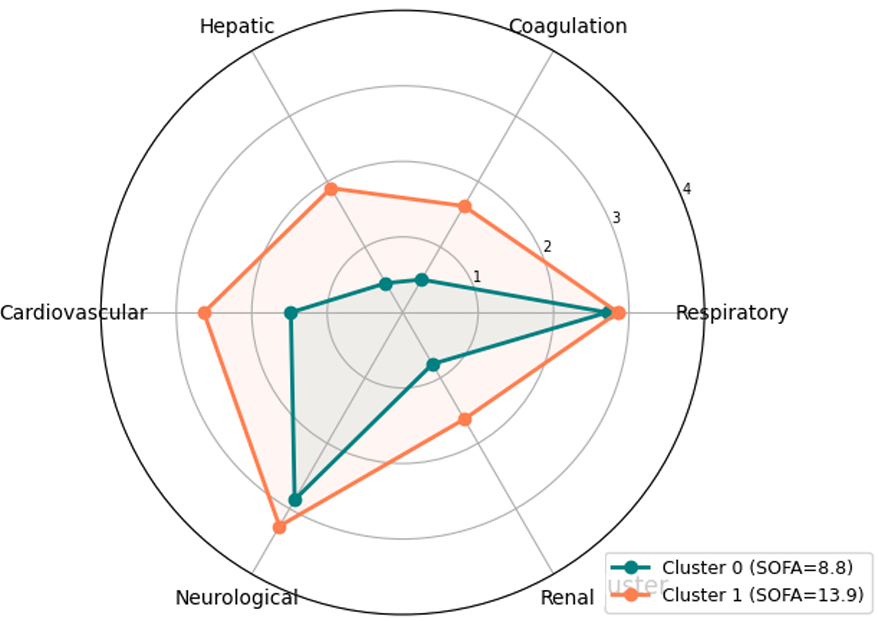}
	\caption{Radar chart comparing mean organ system subscores for Clusters.}
	\label{fig:radar}
\end{figure}
%Radar chart comparing mean organ system subscores for Cluster 0 (improving) and Cluster 1 (persistently severe).

\subsection{SOFA Trajectory Clustering}
\label{sec:cluster-results}

Silhouette score analysis across $k=2$ to $7$ (Appendix~\ref{app:extra}, Figure~\ref{fig:silhouette}) identified $k=2$ as the optimal number of clusters for the 9,488 eligible ICU stays, yielding a moderate silhouette score of 0.456. K-means clustering partitioned these stays into two clinically distinct groups (Figure~\ref{fig:clusters}). Cluster~0, comprising 5,585 stays (58.9\%), exhibited an \emph{improving} trajectory, with a centroid SOFA declining from 9.6 at T1 to 7.5 at T7. Cluster~1, comprising 3,903 stays (41.1\%), followed a \emph{persistently severe} pattern, with a centroid declining only marginally from 14.6 to 13.1, remaining well above the severe band threshold throughout admission.

The radar chart (Figure~\ref{fig:radar}) and Table~\ref{tab:clusters} show that the two clusters were most differentiated by cardiovascular, hepatic, coagulation, and renal involvement, with Cluster~1 exhibiting substantially higher mean subscores across all four. Neurological and respiratory subscores were comparatively similar between clusters, indicating that the persistently severe cluster is distinguished not by worse respiratory or neurological function, but by simultaneous failure across the cardiovascular, hepatic, coagulation, and renal systems. This pattern directly corroborates the component analysis in Section~\ref{sec:component-results}, confirming that the same four organ systems are not only key cross-sectional predictors, but key determinants of whether an ICU stay follows an improving or persistently severe clinical course.

%The radar chart (Figure~\ref{fig:radar}) and Table~\ref{tab:clusters} show that the two clusters were most sharply differentiated by cardiovascular, hepatic, coagulation, and renal involvement, with Cluster~1 exhibiting substantially higher mean subscores on all four systems. Neurological and respiratory subscores were comparatively similar between clusters, indicating that the persistently severe cluster is distinguished not by worse respiratory or neurological function, but by simultaneous failure across the cardiovascular, hepatic, coagulation, and renal systems. This pattern directly corroborates the component contribution analysis in Section~\ref{sec:component-results}, confirming that the same four organ systems are not only the strongest cross-sectional predictors of severity, but also the key determinants of whether an ICU stay follows an improving or persistently severe clinical course.

\section{Discussion and Conclusion}
\label{sec:discussion}

%\textbf{RQ1.} 
The TCN demonstrated moderate predictive accuracy for next-day SOFA scores, confirming that it can extract meaningful signal from ICU physiological data beyond simple persistence, as established by \citet{bednarski2022} for ICU time-series tasks more broadly. However, the SHAP finding that the model anchored predictions almost entirely in the most recent observation, rather than exploiting temporal dependencies, suggests that a three-day input window may be insufficient to leverage TCN advantages in highly autocorrelated clinical settings, ironically exhibiting behaviour closer to the non-sequential approaches of prior SOFA prediction studies~\citep{montomoli2021,asuroglu2021} than to a true sequence model, and limiting its utility as a standalone deterioration detection tool.

\begin{table}
	\centering
	\caption{Mean SOFA organ system subscores.} % by trajectorycluster
	\label{tab:clusters}
	\resizebox{\linewidth}{!}{%
		\begin{tabular}{lcc}
			\toprule
			Organ system & Cluster 0 & Cluster 1 \\
			& (Improving) & (Persist. severe) \\
			\midrule
			Respiratory     & 2.712 & 2.853 \\
			Coagulation     & 0.503 & 1.627 \\
			Hepatic         & 0.448 & 1.901 \\
			Cardiovascular  & 1.482 & 2.635 \\
			Neurological    & 2.864 & 3.267 \\
			Renal           & 0.789 & 1.632 \\
			\midrule
			Mean SOFA total & 8.799 & 13.915 \\
			\bottomrule
	\end{tabular}}
\end{table}

%\textbf{RQ2.} 
Cardiovascular dysfunction was the strongest discriminator of both overall SOFA severity and acute deterioration, consistent with its central role in sepsis-related organ failure, while the finding that respiratory dysfunction behaves differently cross-sectionally versus acutely has direct implications for clinical monitoring: chronic respiratory impairment may be a poor discriminator of overall severity, but acute respiratory changes still warrant close attention during deterioration episodes. The two trajectory clusters (improving and persistently severe) mirror clinically recognised ICU phenotypes, and the convergence of clustering and component analysis on the same four organ systems, cardiovascular, hepatic, coagulation, and renal, suggests coherent biological patterning in organ dysfunction progression that generalises across two independent analytical approaches within this cohort, warranting investigation in larger and more diverse cohorts.

%clinical monitoring prioritisation 

%\textbf{RQ3.} 
%The two trajectory clusters (improving and persistently severe) mirror clinically recognised ICU phenotypes, and the convergence of clustering and component analysis on the same four organ systems, cardiovascular, hepatic, coagulation, and renal, suggests coherent biological patterning in organ dysfunction progression that generalises across two independent analytical approaches within this cohort, warranting investigation in larger and more diverse cohorts.

%\textbf{Conclusions.} 
This study demonstrated that a TCN could predict short-term SOFA score trajectories from routinely collected ICU data with moderate accuracy, while revealing that cardiovascular, hepatic, coagulation, and renal dysfunction collectively govern both cross-sectional severity and longitudinal clinical course in the MIMIC-IV cohort. The complete-case selection approach, while methodologically justified~\citep{websterclark2024}, introduced a selection bias that constrained model performance at low-severity levels and in deterioration detection specifically, highlighting that dataset construction decisions have direct consequences for clinical utility. More broadly, this illustrates that early methodological choices in ICU ML pipelines, particularly around missing data handling, propagate through every subsequent stage of model development and evaluation in ways that are not always apparent until interpretability and stratified analyses are performed. Future work should explore longer input windows, alternative missing-data strategies beyond complete-case selection, and external validation across more diverse ICU populations to establish generalisability beyond a single-centre.

% 2. Bibliography
\printbibliography
%\bibliographystyle{plain}
%\bibliography{references}

% 3. Appendix
\appendix

\section{Supplementary Results, Methodology, and Extended Analyses}
\label{app:extra}

%This appendix reports supporting figures, tables, and detail referenced.

\subsection*{Clinically Stratified and Deterioration-Detection Metrics}
\begin{figure}[h]
	\centering
	\includegraphics[width=\linewidth]{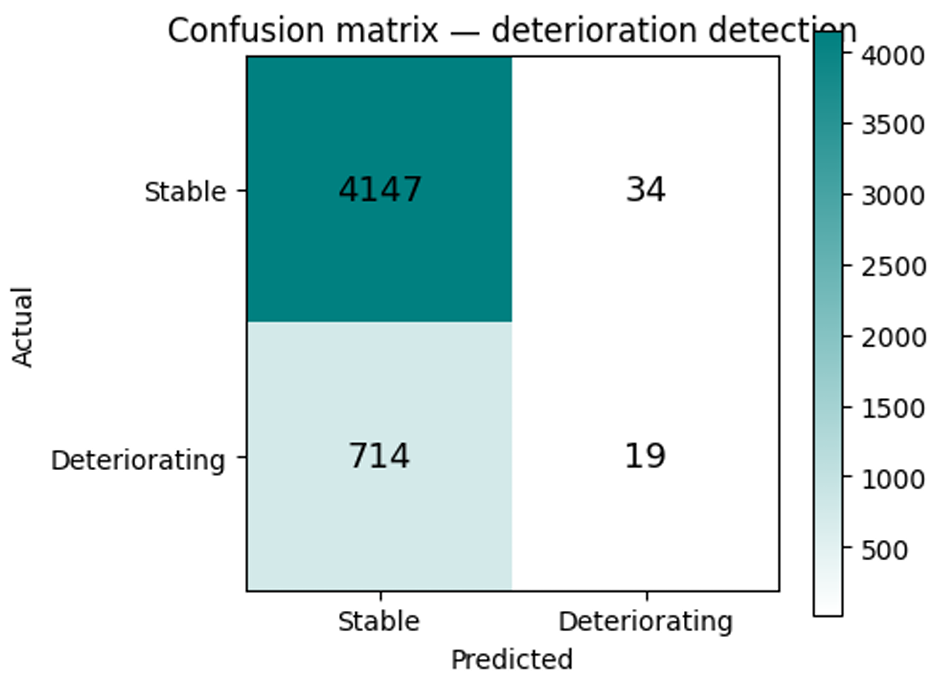}
	\caption{Confusion matrix for deterioration detection (SOFA $\geq$ 2), supporting Table~\ref{tab:deterioration}.}
	\label{fig:confusion}
\end{figure}

This confusion matrix shows the classification outcomes underlying Figure \ref{fig:confusion}. Of 733 truly deteriorating sequences, only 19 were correctly flagged, while 714 were misclassified as stable. Of 4,181 stable sequences, 4,147 were correctly identified and 34 misclassified as 
deteriorating. The asymmetry reflects severe class imbalance pushing the model toward 
predicting stability by default, consistent with the low recall (0.03) and F1-score (0.05) for the 
deteriorating class reported in Table \ref{tab:deterioration}.

\begin{table}
	\centering
	\caption{Classification report for deterioration detection ($\Delta$SOFA $\geq 2$) (supports Section~\ref{sec:regression-results}).}
	\label{tab:deterioration}
	\resizebox{\linewidth}{!}{%
		\begin{tabular}{lcccc}
			\toprule
			& Precision & Recall & F1-score & Support \\
			\midrule
			Stable         & 0.85 & 0.99 & 0.92 & 4{,}181 \\
			Deteriorating  & 0.36 & 0.03 & 0.05 & 733 \\
			\midrule
			Accuracy       & \multicolumn{3}{c}{0.85} & 4{,}914 \\
			Macro avg      & 0.61 & 0.51 & 0.48 & 4{,}914 \\
			Weighted avg   & 0.78 & 0.85 & 0.79 & 4{,}914 \\
			\bottomrule
	\end{tabular}}
\end{table}
Table~\ref{tab:severity} provides the detailed TCN regression performance stratified by SOFA severity band, including the number of sequences and corresponding MAE, RMSE, and $R^2$ values (detail in Section--\ref{sec:regression-results}).

\begin{table}
	\centering
	\caption{TCN regression performance stratified by SOFA severity band (supports Section~\ref{sec:regression-results}).}
	\label{tab:severity}
	\resizebox{\linewidth}{!}{%
		\begin{tabular}{lcccc}
			\toprule
			Band & $n$ (sequences) & MAE & RMSE & $R^2$ \\
			\midrule
			Low (0--5)      & 208   & 1.512 & 1.941 & $-6.256$ \\
			Moderate (6--10)& 1{,}967 & 1.240 & 1.595 & $-0.363$ \\
			Severe ($\geq$11)& 2{,}739 & 1.617 & 2.099 & 0.134 \\
			\bottomrule
	\end{tabular}}
\end{table}

\begin{figure*}[tb]
	\centering
	\includegraphics[width=\linewidth]{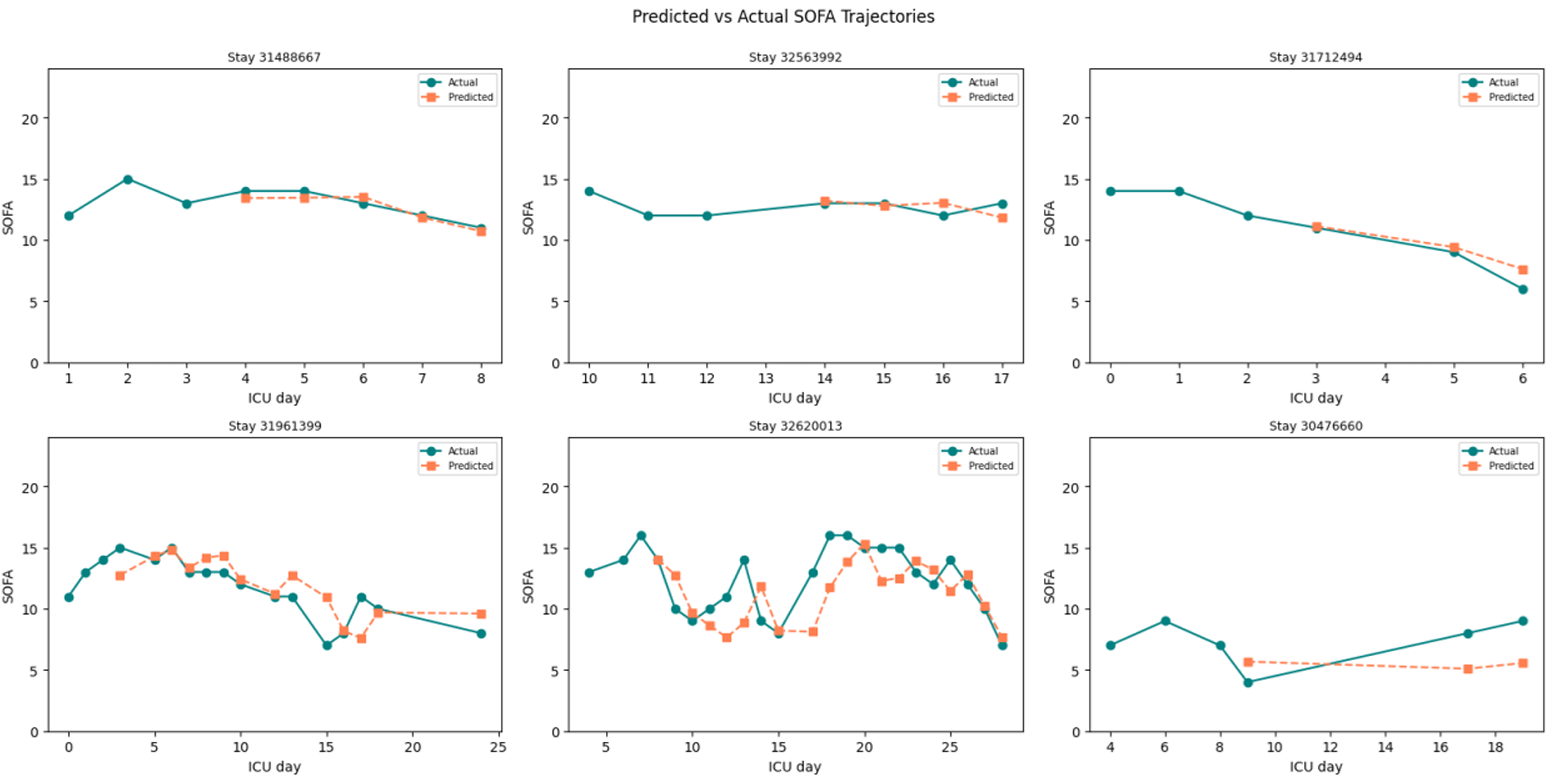}
	\caption{Predicted vs actual SOFA trajectories for six sampled ICU admissions, illustrating the model's smoothing bias under rapid deterioration or recovery.}
	\label{fig:trajectories}
\end{figure*}

\begin{figure*}
	\centering
	\includegraphics[width=\linewidth]{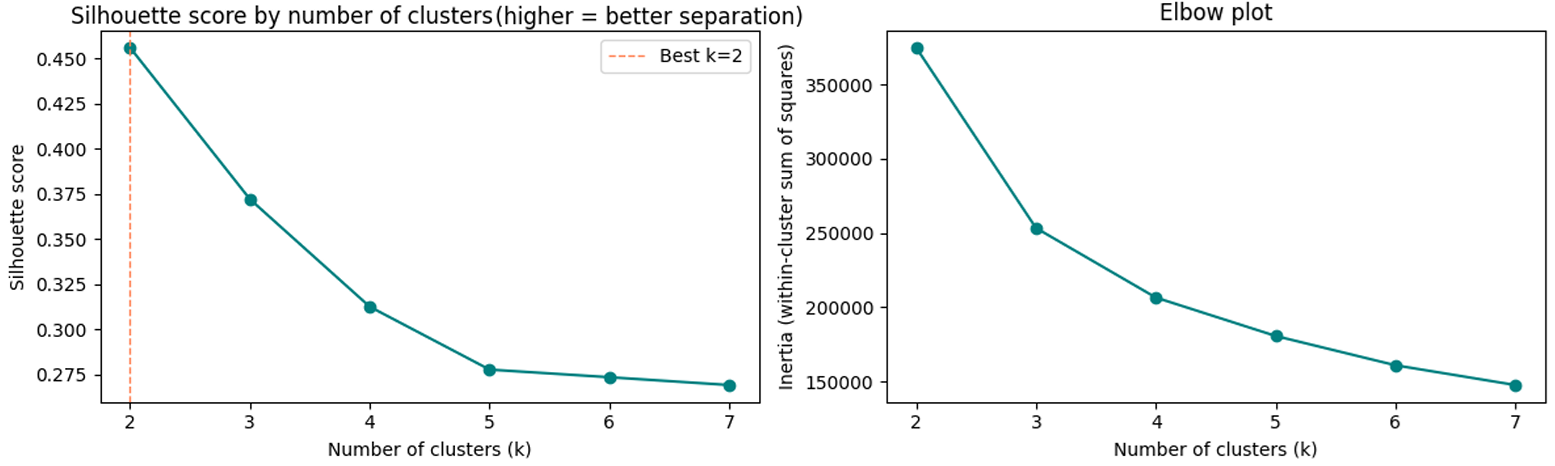}
	\caption{Silhouette score by number of clusters (left) and elbow plot of inertia (right), supporting Section~\ref{sec:cluster-results}.}
	\label{fig:silhouette}
\end{figure*}
Figure~\ref{fig:trajectories} provides representative examples of predicted and observed SOFA trajectories across six ICU admissions, spanning stable, gradual, and volatile clinical courses. Stable trajectories are tracked relatively closely, whereas sharp deteriorations and recoveries are consistently underestimated or missed. In the most volatile admission, deterioration peaks were underestimated by approximately 2--3 points. This tendency to smooth abrupt changes is consistent with the SHAP findings, which indicate that the model places greater weight on the most recent observation than on the preceding trajectory.

%These panels compare predicted and actual SOFA scores across six ICU stays spanning stable, 
%gradual, and volatile clinical courses. Stable trajectories are tracked closely, while sharp 
%deteriorations or recoveries are consistently underestimated or missed. In the most volatile 
%admission, deterioration peaks were underestimated by roughly 2–3 points. This supports the 
%SHAP finding that the model anchors on the most recent observation rather than the preceding 
%trend.
%
%Figure~\ref{fig:trajectories} provides representative examples of predicted and observed SOFA trajectories across six ICU admissions, illustrating the model's tendency to smooth abrupt changes in SOFA.

\subsection*{SOFA Component Contribution}

\begin{figure}[tb]
	\centering
	\includegraphics[width=\linewidth]{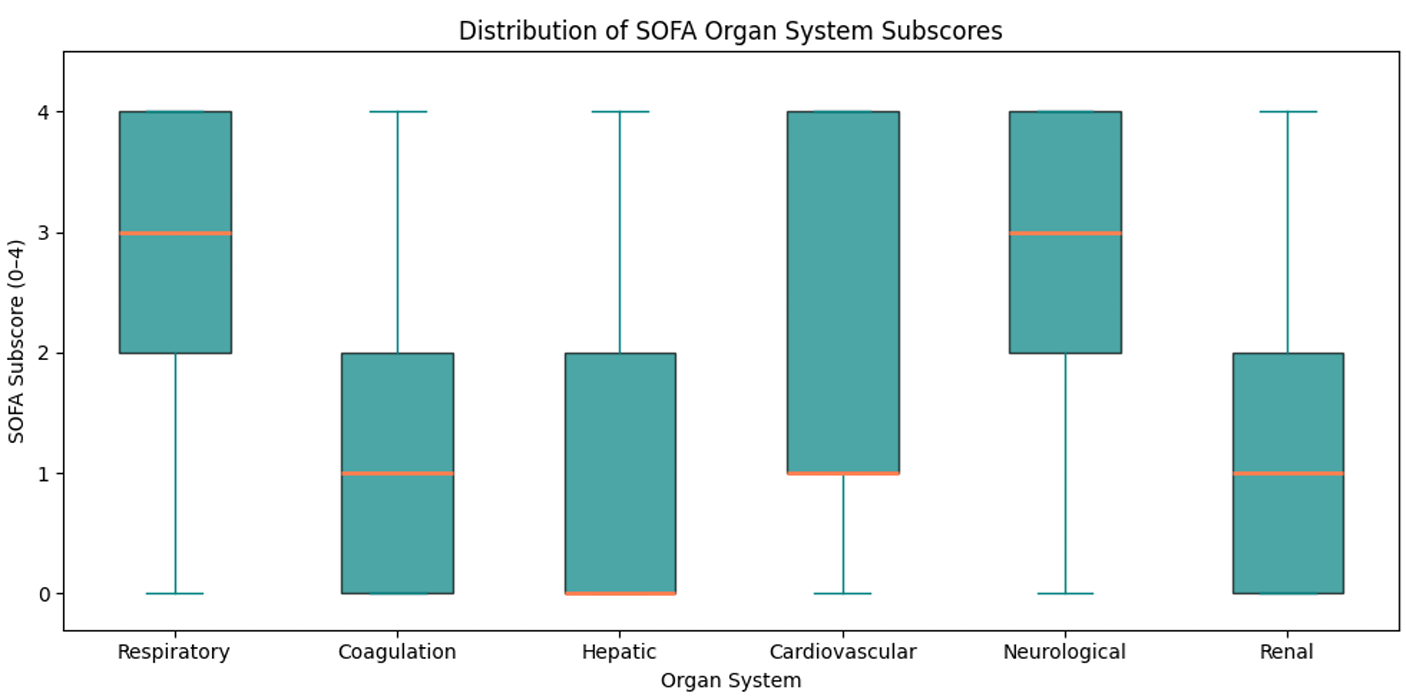}
	\caption{Box plot of SOFA subscore distributions by organ system, supporting Section~\ref{sec:component-results}.}
	\label{fig:boxplot}
\end{figure}

The box plot (Figure~\ref{fig:boxplot}) compares the distribution of each SOFA subscore across the complete-case cohort. Respiratory and Neurological show high medians (3) with narrow spread, reflecting near-universal impairment. Hepatic, Coagulation, and Renal are zero-inflated, with 
dysfunction concentrated in a smaller, severely ill subset. Cardiovascular shows the widest 
spread, consistent with the binary nature of vasopressor use.

\begin{figure}[tb]
	\centering
	\includegraphics[width=0.88\linewidth]{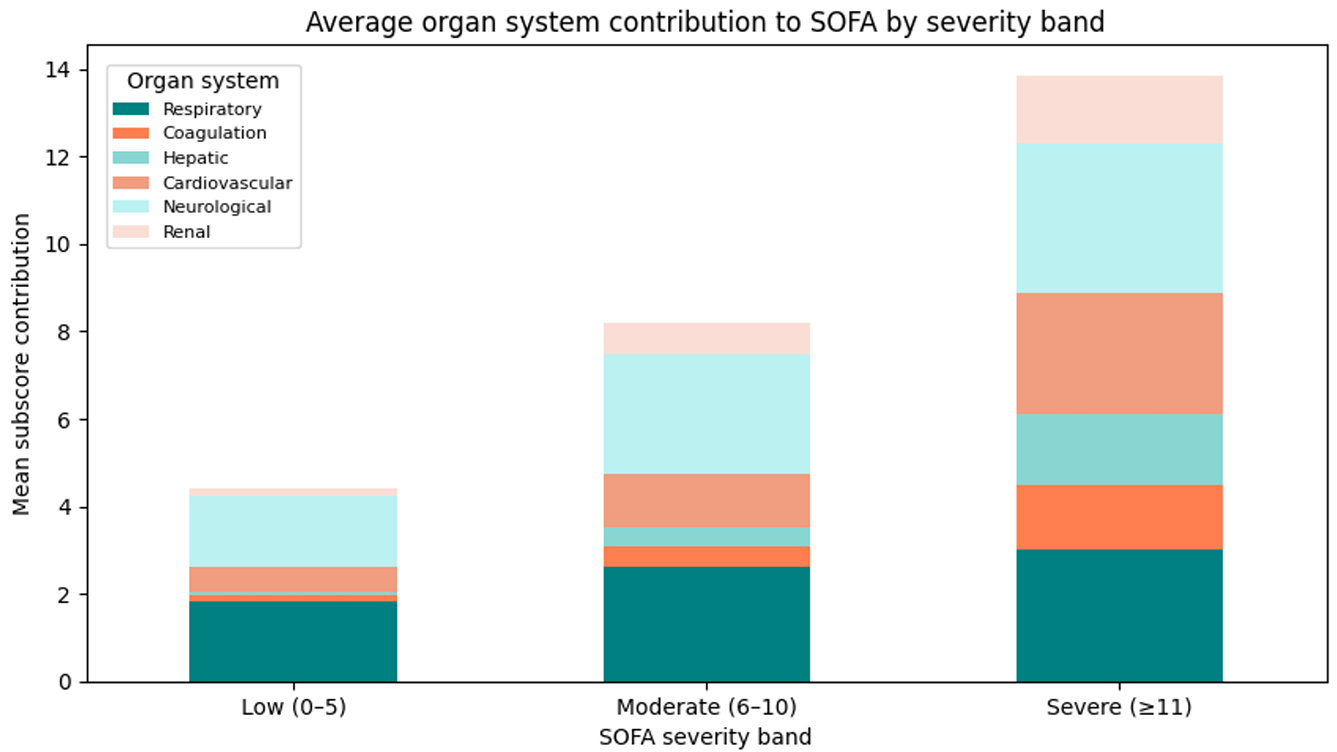}
	\caption{Average organ system contribution to SOFA score by severity band, supporting Section~\ref{sec:component-results}.}
	\label{fig:severity}
\end{figure}

Figure~\ref{fig:severity} (stacked bar chart) shows how each organ system's contribution changes across Low, Moderate, and Severe SOFA bands. Cardiovascular, Hepatic, Coagulation, and Renal rise sharply with severity, with Cardiovascular alone increasing nearly fivefold. Neurological and Respiratory are already elevated even in the Low band, leaving little room for further escalation. This mirrors their low variance contribution reported in Section~\ref{sec:component-results}. 

\begin{figure}[tb]
	\centering
	\includegraphics[width=\linewidth]{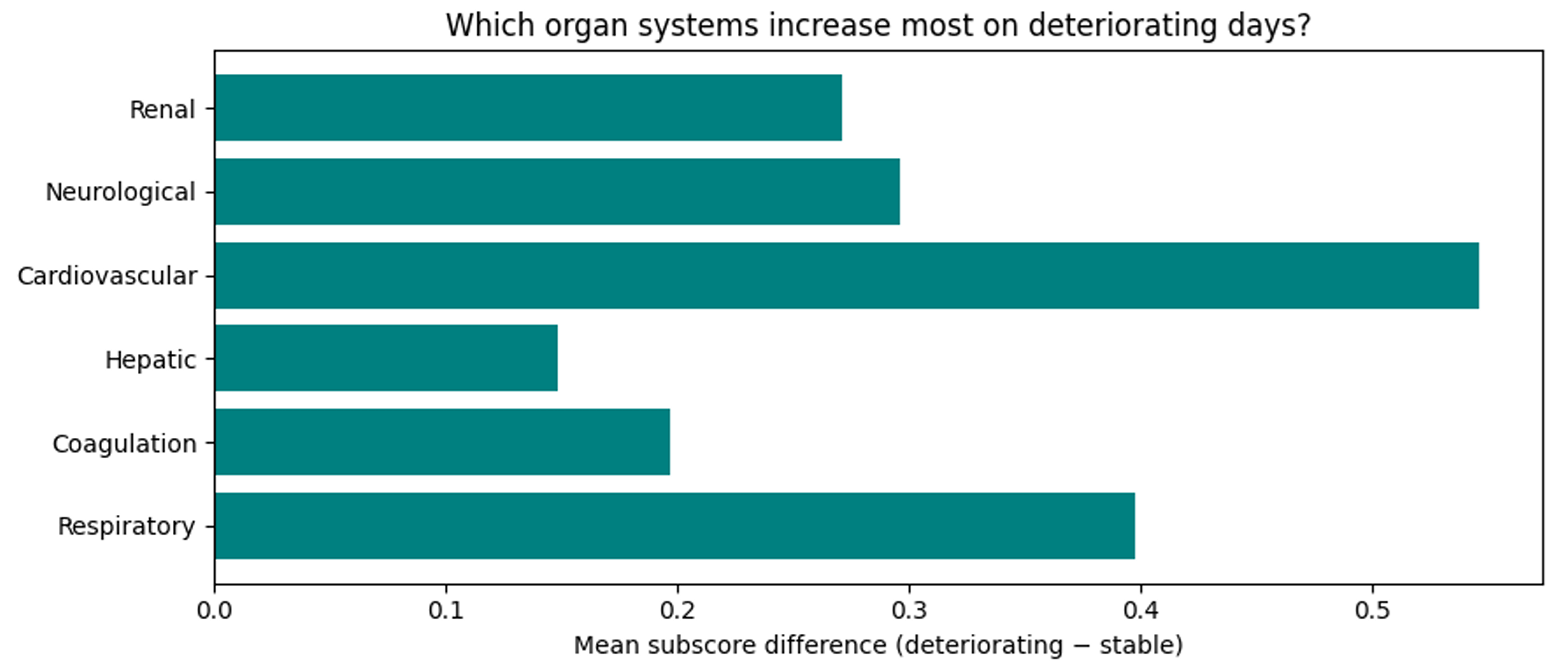}
	\caption{Mean subscore difference between deteriorating and stable days by organ system, supporting Section~\ref{sec:component-results}.}
	\label{fig:deterioration}
\end{figure}

The deterioration graph (Figure~\ref{fig:deterioration}) ranks organ systems by how much their mean subscore rises on deteriorating days ($\Delta$ SOFA $\geq$ 2) versus stable days. Cardiovascular shows by far the largest increase (+0.546), confirming its role in acute clinical worsening. Respiratory ranks second (+0.398) despite ranking last on variance contribution and correlation. This suggests it fluctuates acutely during deterioration despite contributing little to overall severity differences. 

\subsection*{Trajectory Clustering}
The left panel shows silhouette scores (Figure~\ref{fig:silhouette}) across k = 2 to 7, identifying k = 2 as optimal at 0.456. The right panel shows the elbow plot of within-cluster inertia, which declines smoothly with no clear inflection point. Neither diagnostic supports a higher k. Together they support the two cluster solution used in Section~\ref{sec:cluster-results}.

\subsection*{SOFA Variable Extraction Reference}
\label{app:variables}
Table~\ref{tab:variables} summarizes the SOFA organ-system variables used in this study and their corresponding source tables in MIMIC-IV. These variables form the basis of the preprocessing and SOFA score calculation described in Section~\ref{sec:preprocessing}.

\begin{table}[h]
	\centering
	\caption{SOFA organ system variables and their MIMIC-IV source tables, supporting Section~\ref{sec:preprocessing}.}
	\label{tab:variables}
	\resizebox{\linewidth}{!}{%
		\begin{tabular}{ll}
			\toprule
			Organ system & Variables (source tables) \\
			\midrule
			Cardiovascular & MAP, vasopressor rates (chartevents, inputevents) \\
			Respiratory & FiO$_2$, PaO$_2$ (chartevents, labevents) \\
			Neurological & GCS components (chartevents) \\
			Renal & Serum creatinine (labevents) \\
			Hepatic & Total bilirubin (labevents) \\
			Coagulation & Platelet count (labevents) \\
			\bottomrule
	\end{tabular}}
\end{table}

\end{document}